\documentclass[letterpaper, 10 pt, conference]{ieeeconf}  

\IEEEoverridecommandlockouts                              

\usepackage[utf8]{inputenc}
\usepackage{amssymb}
\usepackage{lipsum}
\usepackage{multicol}
\usepackage{graphicx}
\let\labelindent\relax 
\usepackage{enumitem}
\usepackage{amsmath}
\usepackage{multirow}
\usepackage[hyphens]{url}
\usepackage[super]{nth}
\usepackage{hyperref}
\usepackage[normalem]{ulem}
\usepackage{xcolor}
\hypersetup{breaklinks=true}
\newcommand{\add}[1]{\textcolor{black}{#1}}

\newcommand{\RomanNumeralCaps}[1]{\MakeUppercase{\romannumeral #1}}

\usepackage{cite}

\title{\LARGE \bf Caveats on the first-generation da Vinci Research Kit: latent technical constraints and essential calibrations}

\author{Zejian Cui$^{*,{1,2}}$, Jo\~ao Cartucho$^{*,1}$, Stamatia Giannarou$^{1}$, and Ferdinando Rodriguez y Baena$^{1,2}$%
\thanks{$^{*}$These authors contributed equally to the work.}
\thanks{Zejian Cui:
        {\tt\small zejian.cui19@imperial.ac.uk}}%
\thanks{Jo\~ao Cartucho:
        {\tt\small j.cartucho19@imperial.ac.uk}}%
\thanks{$^{1}$The Hamlyn Centre for Robotic Surgery, Imperial College London, London SW7 2AZ, UK}%
\thanks{$^{2}$Mechatronics in Medicine Lab, Department of Mechanical Engineering, Imperial College London, London SW7 2AZ, UK}%
}

\graphicspath{{figures/}}

\begin{document}

\maketitle
\thispagestyle{empty}
\pagestyle{empty}


\begin{abstract}
Telesurgical robotic systems provide a well established form of assistance in the operating theater, with evidence of growing uptake in recent years. Until now, the da Vinci surgical system (Intuitive Surgical Inc, Sunnyvale, California) has been the most widely adopted robot of this kind, with more than 6,700 systems in current clinical use worldwide \cite{daVinci6000}. To accelerate research on robotic-assisted surgery, the retired first-generation da Vinci robots have been redeployed for research use as ``da Vinci Research Kits" (dVRKs), which have been distributed to research institutions around the world to support both training and research in the sector. In the past ten years, a great amount of research on the dVRK has been carried out across a vast range of research topics. During this extensive and distributed process, common technical issues have been identified that are buried deep within the dVRK research and development architecture, and were found to be common among dVRK user feedback, regardless of the breadth and disparity of research directions identified. This paper gathers and analyzes the most significant of these, with a focus on the technical constraints of the first-generation dVRK, which both existing and prospective users should be aware of before embarking onto dVRK-related research. The hope is that this review will aid users in identifying and addressing common limitations of the systems promptly, thus helping to accelerate progress in the field.

\end{abstract}
\section{Introduction}

Robotic-Assisted Minimally Invasive Surgery (RMIS) has gained significant popularity in recent years due to its advantage of causing less tissue trauma and reducing hospitalization time for patients. Of all surgical robotic platforms available in the market, the da Vinci robot (Intuitive Surgical Inc, Sunnyvale, California) has dominated RMIS, with more than 10M operations having been performed on this platform in the past twenty years \cite{dvrk_review_10year}. To further propel RMIS research on the da Vinci surgical platform, in \add{2012}, an initiative was launched to repurpose retired first-generation da Vinci robots by converting them into da Vinci Research Kits (dVRKs) \cite{kazanzides2014open}, which consist of both a software package and hardware controllers. These dVRKs have been distributed to research institutions around the world. Until now, more than 40 research groups across 10 countries have benefited from the initiative, forming a thriving dVRK research community, with more than 250 peer-reviewed research papers on the dVRK having been published. 

\begin{figure}[t]
    \centering
    \includegraphics[width=\columnwidth]{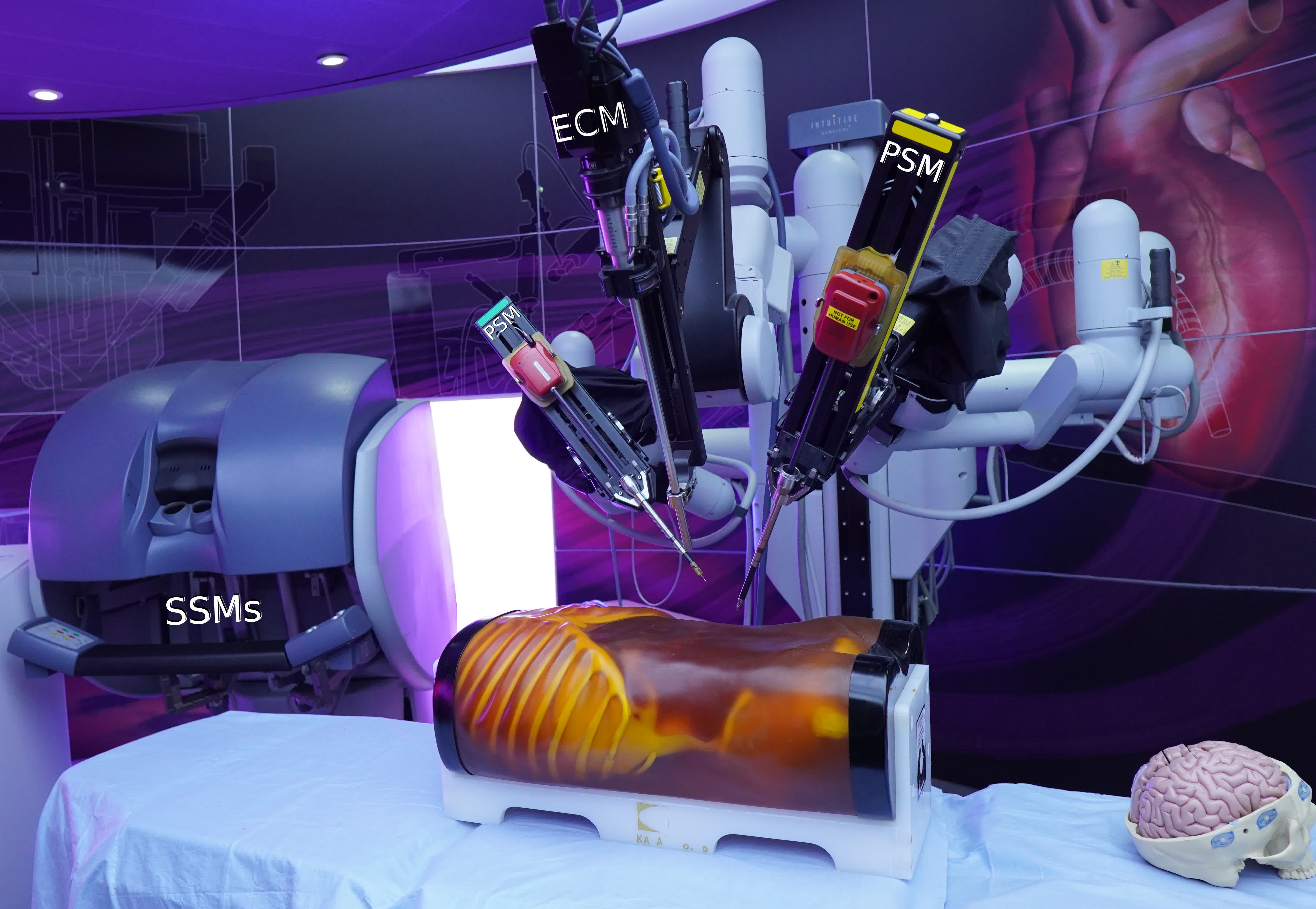}
    \caption{The first-generation dVRK with two surgeon-side manipulators (SSMs), three patient-side manipulators (PSMs) and one endoscopic camera manipulator (ECM)}
    \label{fig:dvrk}
\end{figure}

Although research conducted on the dVRK branches into different topics, almost all of these works deal with data directly provided by the dVRK, containing kinematics, image and system information, as described in \cite{dvrk_review_10year}. These data are arguably the best source of information available to users who wish to keep abreast of the latest state of the robot as they develop novel research on the dVRK. However, blind reliance on the fidelity of these data can be risky. The inaccuracy of dVRK data partly arises from the intrinsic design of the telesurgical platform, and partly from limitations of the factory calibration of the hardware, which has only been improved upon in recent years thanks to a growing interest in the field and the many contributions of researchers working with the dVRK. Awareness of these seminal works should be heightened within the dVRK community, while thus far, there only exists one review paper about the dVRK \cite{dvrk_review_10year}, which focuses on categorizing research in terms of relevant high-level application areas.

To fill in this gap, in this paper we aim to analyze the technical constraints identified within the first-generation dVRK so that users can be aware of these shortcomings and take corrective actions to overcome these, in pursuit of better research outcomes. We have based this review paper on existing literature and the feedback from researchers of the dVRK community, as detailed in Sec. \ref{sec:search_method}. 

According to our initial research, there is a general consensus among dVRK users about the inaccuracies within the dVRK kinematics and image data, which necessitates studies on essential kinematics and camera calibration. Allowing for the fact that hand-eye calibration is an indispensable part of teleoperation, the accuracy of which can have a profound impact on subsequent task implementations, we also stress technical concerns about hand-eye calibration. Additionally, we address potentiometer calibration after weighing up its substantial impact on the performance of autonomous task execution, even though, thus far, to the best of our knowledge, there is no literature about this topic. In the following sections, kinematics calibration, hand-eye calibration, potentiometer calibration and camera calibration problems are detailed in Sec.\ref{sec:Kinematicscalib}, \ref{sec:Hand-eye}, \ref{sec:potentiometer} and \ref{sec:calib}, respectively. For each of these, we first identify the nature of the problem, then distill prevailing schools of thought concerned on how to tackle them, and finally we point out potential paths for better addressing these problems, placed in the dVRK framework. A discussion is presented in Sec.\ref{sec:discussion}, where we summarize the main findings, and other miscellaneous dVRK technical issues uncovered during the survey.
\section{Search Methodology}
\label{sec:search_method}

\subsection{Literature review}

To identify the most significant technical constraints that have hampered research progress with the dVRK, we first looked into all papers on the dVRK across different research application areas and gathered common issues that researchers claimed to have affected task performance, taking advantage of the latest review paper on the subject, which encompasses papers on the dVRK from year 2014 to 2021 \cite{dvrk_review_10year}. We also employed the keywords ``dVRK" paired with ``error", ``calibration" and ``task performance" individually on Google Scholar, ScienceDirect and RefWorks and included all relevant findings in this survey.

\subsection{Consultation with researchers}

\subsubsection{Leading researchers}

Having gathered a list of technical issues reported by dVRK users in the literature, we consulted with leading researchers of the dVRK community (Mr. Anton Deguet and Dr. Simon DiMaio) about whether these issues have been discovered under the first-generation dVRK and thus can be regarded as universal. 

\subsubsection{Peer researchers}

We designed a questionnaire featuring selected technical issues, which we deemed to be ``universal", and had it distributed within the dVRK community\footnote{More info at: \url{https://jhudvrk.slack.com}}.
The design of the questionnaire and a summary of user responses are detailed in Appendix \ref{sec:questionnaire}.

\section{Kinematics Calibration}
\label{sec:Kinematicscalib}
A well-acknowledged technical issue within the dVRK is its time-variant inaccurate forward kinematics, which induces a poor surgical tool tip pose estimation, reflected by a discrepancy in readings of up to 
\add{1.02mm \cite{daVinci_1.02,ferguson2018_1.02}} between the alleged tip pose streamed from the dVRK and the ground truth observed by an external means. This discrepancy, termed positioning error in this paper, has exceeded the sub-millimetre positioning accuracy required by RMIS\cite{baumkircher2022collaborative}, having repercussions on diverse research projects implemented on the dVRK. \add{Note that kinematics errors exist in all individual joints of the dVRK; however, it is in the end effector position that the discrepancy caused by the kinematics errors is most evident, because of the accumulation of errors across all the joints.} 

\subsection{Need for kinematics calibration}
\add{For research on surgical subtask automation, accurate positioning is required, such as in debridement \cite{Berkeley_2_phase_calibration}, suturing \cite{Berkeley_HOUSTON, DualInsertation_Kine_auto}, needle extraction \cite{visual_automated_suturing}, and peg transfer \cite{Berkeley_fiducial_marker, hwang2020superhuman, Berkeley_depthsensing_application}, where a surgical instrument is required to move to the target grasping point.}
For research on the development of Active Constraints (AC, also known as virtual fixtures) via the dVRK, positioning accuracy will affect the proximity query result, and therefore the entire enforcement stage of the AC implementation \cite{Fanny_Kalman_calibration,maria_collision_AC_Hamlyn,Motor_channeling,Task_space_VF}. Even for research on image-based hand-eye calibration \cite{Hand_eye_nonAX_method}, inaccurate surgical tool kinematic data leads to an inaccurate 3D pose estimation and thereby affects the accuracy of its back-projected 2D pixel position, which serves as a criterion for evaluating different hand-eye calibration strategies. 

\subsection{Challenges in accurate positioning}
\subsubsection{Cable-driven effects}
The dVRK surgical platform, which is a cable-driven robotic system \cite{cable_driven_review}, has an inherent issue in obtaining accurate joint readings from the built-in encoders. This is because all encoders are mounted adjacent to actuators but distant from joints, where the actual joint angles are estimated through transmission kinematics \cite{unscented_KF_jointAngle_estimation}. However, unknown parameters such as pulley-cable friction and nonlinearities \cite{Pully_friction_measurement} contribute to inaccuracies in joint angle estimation, which result in dVRK joint encoder readings not reflecting actual joint positions. These errors in joint space subsequently affect the estimated end effector tip pose through Denavit-Hartenberg (DH) parameters and forward kinematics. \add{The DH parameters here only concern active joints.}

\subsubsection{Inaccurate kinematic parameters}
\add{The wear and tear of the dVRK not only induces cable slack \cite{cable_driven_RAVEN}, which affects the parameters in the transmission kinematics, resulting inaccurate joint encoder readings, but also entails instrument damage which leads to additional inaccuracies in the DH parameters of the forward kinematics, causing the estimated tool-tip pose to stray from the actual one.} \add{Another significant source of error is due to the potentiometers, as explained in detail in \ref{sec:potentiometer}.}
\subsubsection{Other non-kinematic error sources}
\begin{figure}
    \centering
    \includegraphics[width=0.7\columnwidth]{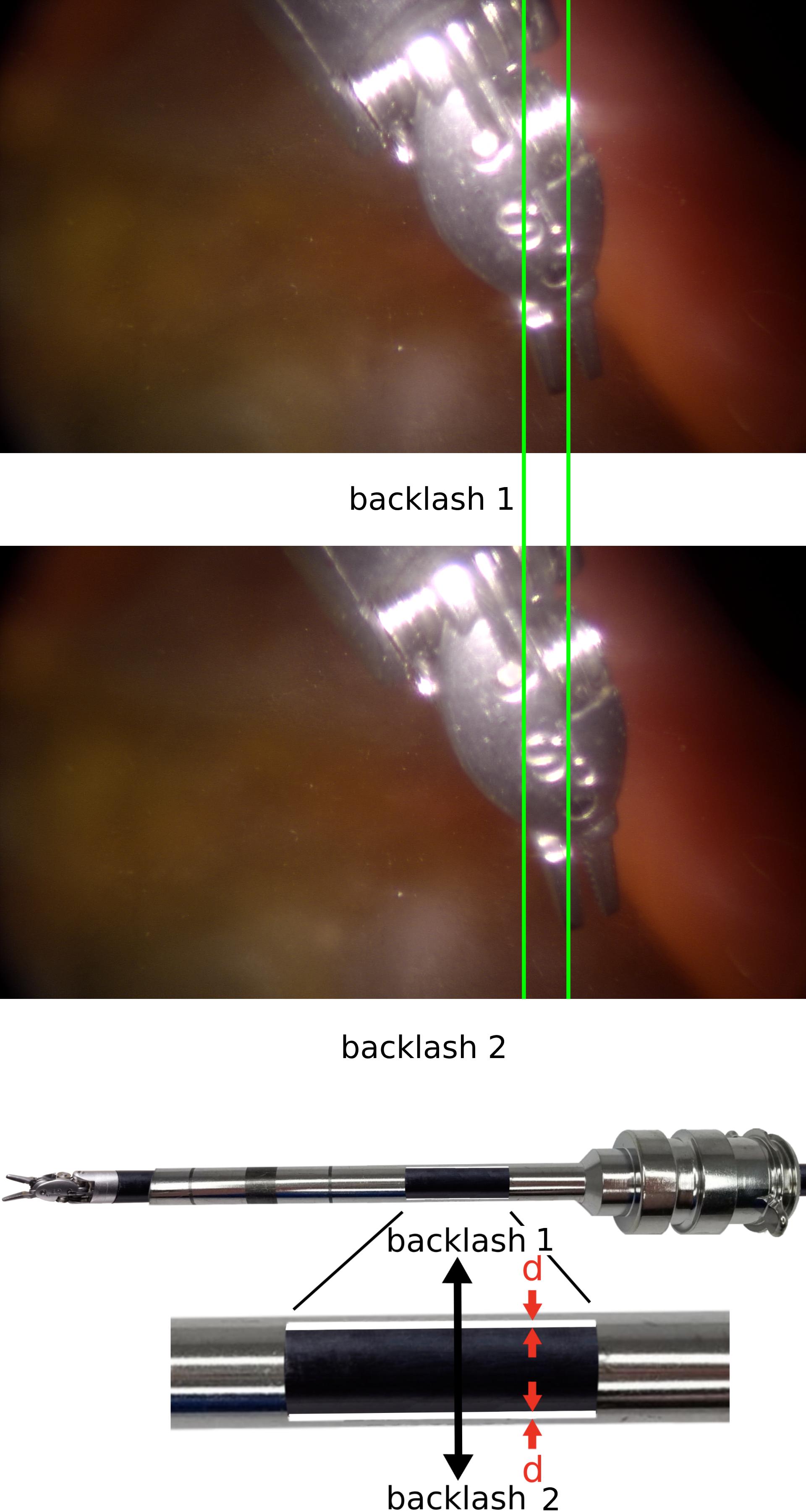}
    \caption{As illustrated in red (d), there is a gap between the surgical instrument's shaft and the cannulae. Therefore, the point of contact between this shaft and the cannulae can change. This phenomenon is known as backlash. Backlash causes the tool to move without being reflected by the encoder readings. Backlash is evident in the figure, illustrated in green, where the ``S" character of Intuitive's logo has moved to a different location, given that the camera is static.
    }
    \label{fig:backlash}
\end{figure}

\begin{figure}
    \centering
    \includegraphics[width=0.9\columnwidth]{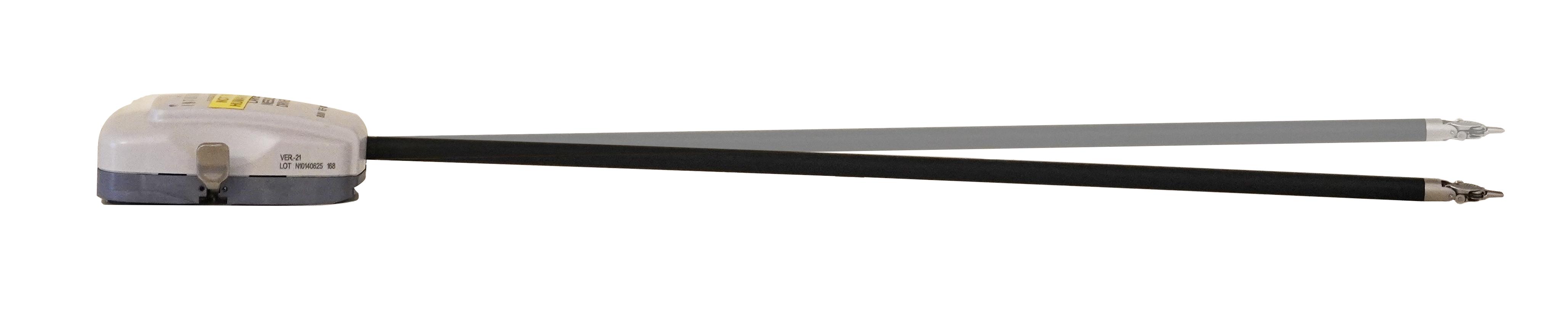}
    \caption{Compliance of the surgical instrument. Similarly to backlash, this compliance is not reflected on encoder readings, and hence not reflected in the joint position estimation.
    }
    \label{fig:deflection}
\end{figure}

Other than kinematic factors, non-kinematic factors such as external forces exerted onto the tool shaft and backlash between the tool shaft and cannula, illustrated in Fig. \ref{fig:backlash}, would also contribute to an overall erroneous end-tip pose estimation \cite{dVRK_compliance_model,dVRK_kinematics_calibration_thesis}. \add{These external forces will affect the tension of cables associated to tool tips, leading to an error in transmission kinematics. In addition, they will also cause a displacement of the tool shaft because of compliance, as illustrated in Fig. \ref{fig:deflection}; this displacement is not detectable by the encoders \cite{dVRK_compliance_model}, and thus positioning error accumulates.}

\subsection{Methods for kinematics calibration}
In this subsection, we gather methods that are available in the current literature to compensate for dVRK positioning errors, and we classify these in term of the specific error source they are aiming to address, namely encoder readings, kinematics parameters and other non-kinematic factors. \add{Methods within the former category aim to estimate the current joint positions, whereas those within the latter category aim to force the system to reach the desired position.}
\subsubsection{Compensation for encoder readings}
For methods that aim to compensate encoder reading inaccuracies, we further divide these into two classes: encoder reading calibration and tool-tip pose estimation. Both classes recognize the nature of the erroneous encoder readings; however, the former focuses on finding actual joint positions in joint space, while the latter places an emphasis on finding the actual tool-tip pose.
\paragraph{Encoder reading calibration}
Methods for calibrating encoder readings follow a standardized pipeline that involves modeling, measurement and parameter fitting \cite{robot_calibration_overview}. To start with, a model parameterized by $\eta$, which maps encoder readings obtained from transducers to actual joint positions, is conceived. In the subsequent measurement stage, robot joints are commanded to move to designated positions while ground truth values of joint positions are gathered via an external measurement system. Finally, different fitting models, which can be either linear or nonlinear, are adopted to estimate the $\eta$ that best fits designated joint positions to ground truth measurements.
 
Major differences between methods in this class are the selection of calibration models and approaches to measuring the ground truth. Following the calibration paradigm, Huang \textit{et al.} \cite{dVRK_kinematics_calibration_thesis} started off using a linear model to map encoder readings to ground truth values gathered by tracking infrared markers mounted on the end effector with an optical camera system, namely the NDI Polaris \cite{NDI_Polaris}. To better estimate and account for the nonlinear nature of cable stretch and friction, Hwang \textit{et al.} adopted a data-driven approach to training deep learning models that map collected readings to actual joint positions gathered in a similar way \cite{Berkeley_fiducial_marker}.

With access to the forward kinematics of a system, which maps encoder readings to actual joint positions, one can find an inverse calibration model that generates encoder commands that drive a robot to reach desired joint positions. Such methods lend themselves to implementing automatic surgical tasks \cite{Berkeley_2_phase_calibration,Berkeley_depthsensing_application,Berkeley_fiducial_marker}, where a surgical tool is requested to accurately move to a given pose. \add{However, there are several drawbacks in model-based calibration methods.} \add{First,} as \add{pointed out} in \cite{UCSD_unifed_2022_TRO}, model-based calibration methods are hard to implement outside of a lab setting because of the need for additional sensors, \add{calibration objects}, and tedious data collection procedures. 
\add{Second}, an invariant calibration model fails to accurately reflect a dynamic mapping relationship as, e.g., the cable wears and tears over time. 
\add{Third, adopting model-based approaches requires greater efforts to find the ground truth values of the joint positions for the 7-DoF da Vinci robot. Although there exists numerical inverse kinematics solvers for the SSMs and the PSMs, running at 1.5 KHz \footnote{One numerical solver is provided at \url{https://github.com/jhu-cisst/cisst}}, there is a lack of kinematics parameters identification procedures. Hence the fidelity of the nominal kinematics parameters for the disposable surgical instruments relies on the quality of manufacture, which, however, can be imprecise.} 
\add{Furthermore, the accuracy of the ground truth measurement provided by an external sensor, a depth-sensing camera for example, is also inevitably affected by the inaccuracy in camera parameters and the hand-eye transformation matrix, as analyzed in \ref{sec:calib} and \ref{sec:Hand-eye}, respectively.}

\paragraph{Tool-tip pose estimation}
 In addition to generating joint commands that drive a robot to reach the desired state, there are times when the actual tool-tip pose in the Cartesian space is more important, not least in situations where the minimum distance between a surgical tool and its surrounding geometry needs to be precisely returned. Typically, pose estimation methods utilize both kinematic and image data. First, a tool-tip pose inferred from encoder readings and the forward kinematics of the system serves as an initial guess, then \add{another} tip-pose is obtained out of \add{2D} image data, serving as an observation. By combining the initial guess with the observed result, a more accurate tip-pose can be estimated. 
 
 Literature on dVRK-based tip-pose estimation varies mainly in terms of the choice of computer vision-based strategy for 3D tool-tip \add{pose reconstruction} and the selection of 3D pose estimators. Moccia \textit{et al.} \cite{Fanny_Kalman_calibration} proposed a binary segmentation-based 3D pose reconstruction strategy, where the 3D pose is reconstructed by first finding the tool-tip on both binary-masked images captured by a stereoscopic endoscopic camera, followed by triangulation. Without relying on a pre-trained tool segmentation model, Richter \textit{et al.} \cite{UCSD_unifed_2022_TRO} simply attached markers on tool joints and used these to correct the discrepancy between detected markers and back-projected joint positions on the image plane. \add{Ye \textit{et.al.} \cite{Menglong_toolPoseEstimation} leveraged the CAD model of surgical instruments and conducted keypoints matching between online generated part-based templates and 2D images to measure the accuracy of the tool pose estimated from the dVRK.} For 3D pose estimators, Extended Kalman Filters (EKFs) and Particle Filters have also been proposed \cite{Fanny_Kalman_calibration,UCSD_unifed_2022_TRO,UCSD_Kalman_augmented,Menglong_toolPoseEstimation}.
 
 Given that most pose estimation methods exploit image data, the accuracy of 3D tool tip pose reconstruction is highly reliant on the quality of gathered images and the robustness of the \add{pose reconstruction} strategies. Therefore, image-dependent pose estimation methods are likely to deliver subpar performances in situations when the camera fails to capture surgical tools in full view or when blood-stained surgical tools fail to be detected and segmented, which are likely to occur within a surgical setting. In addition, the accuracy of image-based methods is also dependent on the precision of camera calibration methods, which are elaborated in Sec. \ref{sec:calib} 
 
 \add{In theory, estimator-based pose estimation methods can also be utilized for estimating joint positions $j_1$-$j_6$; however, as suggested by the findings in \cite{UCSD_unifed_2022_TRO}, an increase in the number of unknown variables increases the system nonlinearity and yields less accurate estimation results. It is also worth mentioning that spurious feature detection and false feature correspondences could also impair the estimation accuracy. Finally, the computational efforts for feature detection in some pose estimation methods are not negligible, such as edge detection in \cite{UCSD_unifed_2022_TRO}, tool segmentation in \cite{Fanny_Kalman_calibration}, which may render these methods less suitable to be incorporated in a modern clinical setting.}
 
\subsubsection{Compensation of kinematics parameters}
The dVRK returns the surgical tool-tip pose based on both joint readings and the system's forward kinematics, which means that calibrated encoder readings are a necessary but not sufficient condition for accurate tool tracking. The nominal dVRK DH kinematics parameters assume that adjacent joint axes are orthogonal and without intersections \cite{Fanny_dvrk_modelling_dynamics}; however, these assumptions fail to hold up in a real clinical setting. In \cite{dVRK_kinematics_calibration_thesis}, authors have shown that there are intersections between adjacent tool axes and additional calibration needs to be conducted on the dVRK kinematics parameters to account for these. A standardized approach to calibrating these kinematics parameters relies on external sensors attached onto the joints, which are used to track complex movements as the robot is driven along random trajectories. This process enables joint axes vectors to be estimated, from which corresponding kinematics parameters can be extracted using linear algebra. 
\subsubsection{Compensation of \add{external force} factors}
Research on this topic stems from the observation that the dVRK surgical tools are intrinsically mildly compliant and can deflect under external environmental forces. However, such deflection is undetectable by joint encoder readings, and hence the end tip pose estimation can suffer further. \add{One way to compensate for the external force factors is through estimation of the external forces/torques.} In \cite{dVRK_compliance_model}, it is assumed that the deflection on the slender dVRK tool shaft is the most salient one, and a compliance model is established by connecting the deflection of the surgical tool in the lateral direction with encoder readings for the first two joints. \add{Alternatively, Yilmaz \textit{et.al.} \cite{Force_factor_2_Jieying} proposed a neural network-based method for estimating inverse dynamic torques such that the external forces/torques can be obtained by subtracting the inverse dynamic torques from the joint torque readings. In both cases, the estimation of the external force factors builds upon the knowledge of system dynamic equations, which are supported by studies on dVRK dynamics modeling, such as \cite{Force_factor_1_wang2019convex}. Another way to address the external force factors is to directly monitor the forces/torques by attaching force sensing devices without further modifications of the original instruments, as proposed in \cite{force_feedback_wpi_thesis}, although there is still great room for improvement in sensing accuracy.}

Apart from joint-level calibration approaches, it is also possible to interpret and account for positioning errors through hand-eye calibration, as seen in \cite{zhangLin_handeyecalib,UCSD_unifed_2022_TRO,Menglong_toolPoseEstimation}, where the positioning error is compensated by conducting hand-eye calibration on the fly. Methods falling within this category are discussed in Sec. \ref{sec:Hand-eye}. 
\section{Potentiometer Calibration}
\label{sec:potentiometer}

\add{Potentiometer calibration is part of the kinematics calibration process since the potentiometers are used to calculate the joint encoder readings, which serve as input to the kinematic parameters. In this paper, we present it as a separate section since we explore potentiometer calibration in detail.} 
To the best of our knowledge, so far there is no explicit mention of potentiometer calibration in the context of dVRK literature.

 The dVRKs are conjured out of the retired first-generation da Vinci systems which were previously used for commercial purposes. The original potentiometer calibration values, for the commercial version of the system, are not always available, and even when available, the calibration values were obtained using a different reference Voltage. This reference Voltage is required to get digital readings, using analogue-to-digital converters (ADCs), since the first-generation da Vinci systems use analogue potentiometers \cite{kazanzides2014open}. The ADCs and other electronic components in the dVRK are different from those employed on the commercial version of the system, and therefore, a different reference Voltage is required. Additionally, the potentiometer calibration values tend to change over time, and hence recalibration is necessary.

All manipulators of the dVRK robot consist of a series of individual joints that are controlled by actuators. Each actuator is coupled with an encoder (directly attached to the actuator) and a potentiometer (indirectly attached to the actuator), both of which can be used to measure the actuator position ($\beta$), which can be further converted to the joint encoder reading by multiplying with given coupling matrices and gear ratios.
According to Walker \textit{et al.} \cite{walker1990getting}, the actuator position ($\beta$) can be returned from both the potentiometer and the encoder using the following equations:

\begin{equation}
   \beta_P = k_P \times P + b_P
   \label{eq:pot}
\end{equation}

\begin{equation}
   \beta_E = k_E \times E + b_E
   \label{eq:enc}
\end{equation}

\begin{equation}
   \beta = \beta_E \approx \beta_P
   \label{eq:joint_state}
\end{equation}
where $P$ is the Voltage reading from the ADC converter of the potentiometer; $k_P$ is a scale factor; $b_P$ is an offset for the Voltage readings; $E$ represents the reading from the encoder counter; $k_E$ is the radians per encoder count, which is known a priori; and $b_E$ is an offset for the encoder.

 For the first-generation dVRK, actuator positions measured from the potentiometer $\beta_P$ are usually noisier than the ones measured from the encoder $\beta_E$, and hence the robot selects $\beta = \beta_E$. Note that $b_E$ needs to be accurately calibrated for each actuator, to ensure an accurate tool-tip pose estimation result. As explained next, $b_E$ is only accurate if $k_P$ and $b_P$ are also accurate, hence the need for potentiometer calibration.
 
\subsection{Need for potentiometer calibration}

\subsubsection{Encoder calibration}

The first-generation dVRK robots use relative encoders, which require calibration of $b_E$ every time the dVRK controllers are powered ON, via the equation below (resulting from Eqs. \ref{eq:pot}, \ref{eq:enc}, \ref{eq:joint_state}):
\begin{equation}
    b_E = -k_E \times E + k_P \times P + b_P
    \label{eq:calib_enc}
\end{equation}

It is shown that the accuracy of $b_E$ depends on the accuracy of $k_P$ and $b_P$, which are the outputs of the potentiometer calibration. This explains why appropriate potentiometer calibration is an important prerequisite towards obtaining an accurate value of $b_E$. 

\subsubsection{Safety}

With the assumption that $\beta_E \approx \beta_P$, the dVRK potentiometers work as redundant sensors in tandem with their encoders to ensure task safety. In the unlikely event that the encoders fail to reflect the actual actuator positions, which may cause the manipulator to keep moving towards an unachievable target position, the discrepancy $| \beta_E - \beta_P |$ can be used to monitor whether the encoders are functional and to decide whether the robot needs to be stopped in order to avoid potential damage. For this, accurate potentiometer calibration is needed.

\subsection{Challenges in potentiometer calibration} \label{sec:calib_kp}
 The goal of potentiometer calibration is to accurately estimate $k_P$ and $b_P$. The scale factor $k_P$ of a potentiometer is calibrated by measuring $E$ and $P$ at different time steps ($t_1$ and $t_2$), during which a large actuator motion is enforced, as follows:

\begin{equation}
    k_P = k_E \frac{E_{t_2} - E_{t_1}}{P_{t_2} - P_{t_1}}
\end{equation}

\add{However, it should be noted that the assumption of a constant value for $k_P$ may not hold true in practice, as real-world potentiometers have inherent non-linearity (i.e., Eq. \ref{eq:pot} assumes linearity). Potentiometer non-linearity refers to the deviation of the potentiometer's output from a linear relationship with the input rotation of the potentiometer's shaft, along its range of motion. This is due to manufacturing errors, physical wear etc.}

Besides $k_P$, the offset $b_P$ of a potentiometer is hard to estimate. It is required to place the actuator at the zero-position, where $\beta_P = 0$ so that we have $b_P = - k_P \times P$, via substitution in Eq. \ref{eq:pot}. 
The challenge here is that it is hard to determine the exact zero-position of an actuator, especially in the context of this research system.

\subsection{Methods for potentiometer calibration}

\add{The methods described in this section can be implemented preoperatively.}

\subsubsection{\add{Potentiometer nonlinearity calibration}}

\add{Common solutions for correcting potentiometer nonlinearity are based on Bezier or polynomial models \cite{song2022research}. These models are used to create a mathematical function that can correct for non-linearities in the potentiometer output. Another approach is to create a lookup table. This involves measuring the output of the potentiometer with its shaft rotated to various input angles and creating a lookup table that maps input angles to corrected output values. Different types of interpolation (e.g., linear, or spline) can be used to estimate corrected output values for input angles that were not measured directly. The lookup approach is also generally faster since it reduces the need for complex calculations.
}

\subsubsection{Calibrating potentiometer offsets ($b_P$) on a PSM}

Each PSM consists of a total of seven actuators, hence there are seven potentiometer offsets that need to be calibrated. The offsets for the last four potentiometers ($b_{P4}$, $b_{P5}$, $b_{P6}$, $b_{P7}$) are usually estimated first with a simple process, whereas calibrations of the offsets $b_{P1}$, $b_{P2}$ and $b_{P3}$ are harder to implement but have a more significant impact on the accuracy of tip-pose estimation, since the first three actuators are connected to the largest links of the PSM.

The solution for calibrating $b_{P4}$ to $b_{P7}$ is to attach a rigid holder to the four wheels that control the movement of a surgical tool. When the wheels are connected to the holder, the last four actuators are held at the zero-position, where $b_P = - k_P \times P$. The holder can either be 3D printed or laser cut, using models that are available on the dVRK's Wiki \cite{dvrk_wiki_potentiometer}.

\begin{figure}
    \centering
    \includegraphics[width=\columnwidth]{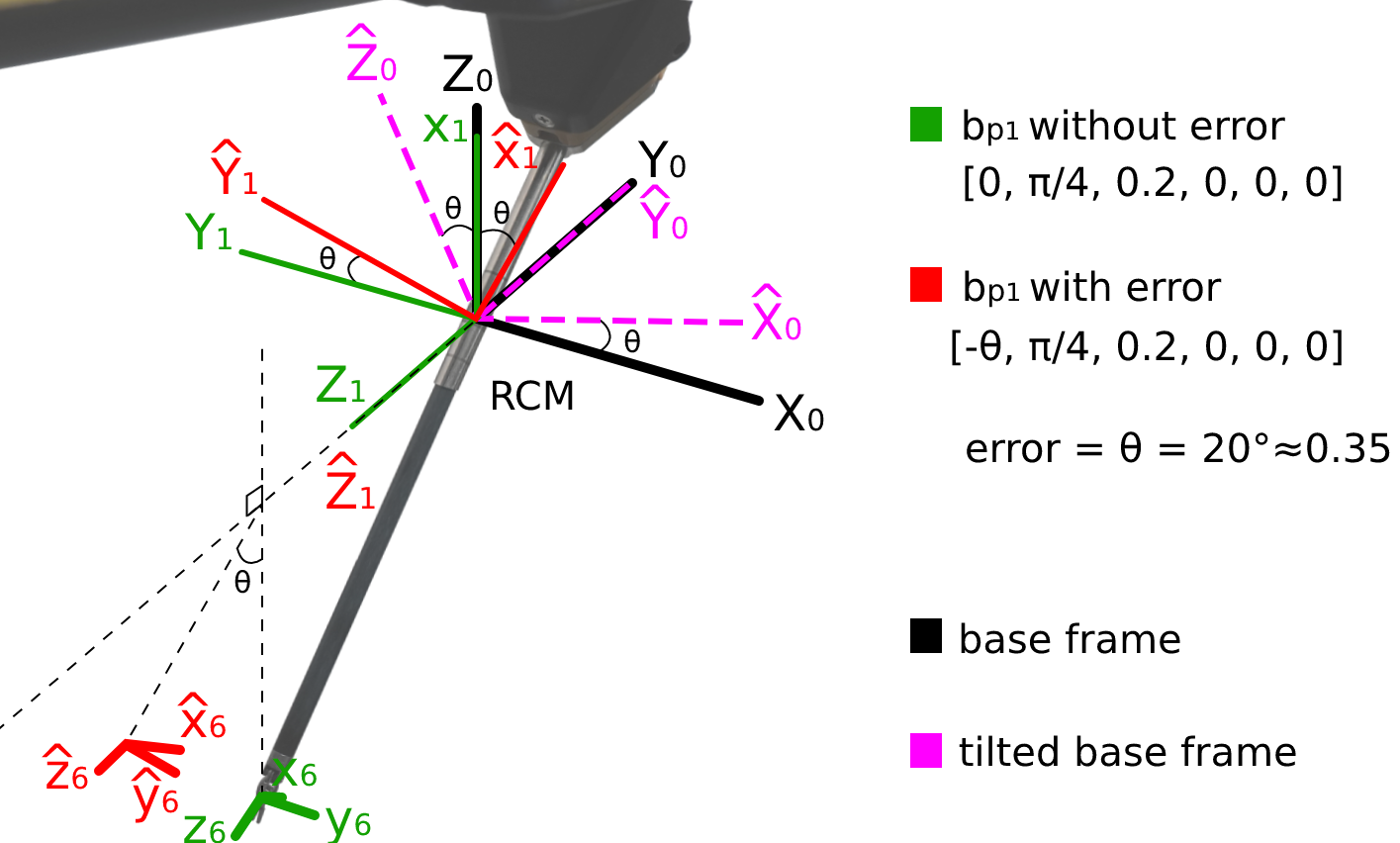}
    \caption{Illustration of the impact of the offset error ($b_{P1}$) in the potentiometer of the first joint of the PSM. For the sake of illustration, an error of 20\textdegree, 0.35 \textit{rad}, is added onto the first joint of the PSM (\^{1}), causing a rotation of 20\textdegree at the tool-tip position (\^{6}). With this error, hand-eye calibration results in a transformation from the camera frame to the tilted base frame illustrated in pink (\^{0}). Hand-eye calibration rectifies the error $b_{P1}$ because the erroneous rotation is compensated by the tilted frame, which is rotated in the opposite direction of the error. Note that joint 3 is in \textit{m} instead of \textit{rad}.}
    \label{fig:pot_offset}
\end{figure}

As illustrated \add{in red} in Figure \ref{fig:pot_offset}, regarding the \nth{1} potentiometer of the PSM, a constant error in the offset (a wrong $b_{P1}$ value after powering ON the PSM), would cause a constant rotation error around the \nth{1} joint. \add{A mathematical description of this effect is the next sub-section (Sec. \ref{sec:offset_compensation})}. The tool-tip pose would suffer from the same rotation error. Here, we assume that the scale ($k_{P1}$) has been correctly calibrated as explained in Sec. \ref{sec:calib_kp}. If the user's goal is to obtain the tool-tip pose relative to the camera, then the user usually needs to perform hand-eye calibration after powering ON the PSM. This hand-eye calibration, in the presence of an offset error in $b_{P1}$, would result in a transformation from the camera to a tilted base frame, shown in pink in Figure \ref{fig:pot_offset}. Errors in calibration of offsets $b_{P1}$ introduce a constant error in the joint positions (since $b_{E1}$ in Equation \ref{eq:calib_enc} depend on $b_{P1}$). Therefore, the tool-tip cartesian pose relative to the PSM's base will be erroneous. However, if hand-eye calibration is used, the tool-tip cartesian pose relative to the camera will be correct, since the hand-eye calibration has already compensated for the errors by outputting a transformation from the camera frame to a tilted PSM base frame.

\add{The compensation approach mentioned above does not apply to the \nth{2} joint or the \nth{3} joint, as further detailed in the next sub-section (Sec. \ref{sec:offset_compensation}). The reason is that no constant transformation matrix exists when the potentiometer offsets ($b_{P2}$ and $b_{P3}$) are not well calibrated.} \add{To quantify the impact of the offset error $e_2$, resulting from an uncalibrated second potentiometer, we generated a reference trajectory for the end-effector with a total of 1,000 points, by interpolation of joint positions ranging between $[-\frac{\pi}{5}, -\frac{\pi}{6}, 0.2m, 0, 0, 0, 0]$ and $[\frac{\pi}{5}, \frac{\pi}{6}, 0.2m, 0, 0, 0, 0]$. We then repeated generating the same trajectory but added $e_2$ to the second joint position to simulate an end-effector trajectory under the presence of an erroneous $b_{P2}$. We found that each 1\textdegree increment in $e_2$ resulted in an additional RMSE of $3.4 \pm 0.1mm$ between the reference and the actual trajectory. This error was observed on surgical instruments including Large Needle Driver (400006), ProGrasp Forceps (400093), Cadière Forceps (400049) and Maryland Bipolar Forceps (400172). If $e_2$ reaches 3\textdegree, the discrepancy in the end effector trajectory will be slightly over 1 $cm$, which highlights the need for calibration of the second potentiometer offset.} Currently, there is no existing technique for accurate calibration of the second potentiometer's offset ($b_{P2}$).

Conversely, a recent technique has been proposed \cite{dvrk_wiki_potentiometer_3rd_joint} to calibrate $b_{P3}$. The general idea of this technique is to place a reference point of the kinematic chain, that is easy to visually track, at the remote centre of motion (RCM). If $b_{P3}$ is accurately calibrated, then this point should not move when located on the RCM, independently of joint 1 and joint 2 actuator positions. Specifically, this technique uses as a reference point the axis between the coordinate frame of joint 4 ($O_{4}$) and joint 5 ($O_{5}$), which is located at the beginning of a surgical instrument's wrist. If a camera is recording this reference point, then its pixel position should remain static. Therefore, if the reference point moves when the position of joint 1 or joint 2 changes, then it means that the offset ($b_{P3}$) is still not accurately calibrated. This technique tries different insertion positions until the motion of the reference point is minimized when the reference point is closest to the RCM. Once the insertion position is known, $b_{P3}$ can be directly measured. \add{The results of our questionnaire indicate that 60\% of the respondents were not aware of this technique used to calibrate the third potentiometer offset.} 

\subsection{Compensation for potentiometer offset errors using hand-eye calibration}
\label{sec:offset_compensation}

\add{As previously mentioned, potentiometer offset errors  result in deviations between the actual joint frames ($0$, $1$, and $2$) and the virtual frames ($\hat{0}$, $\hat{1}$, and $\hat{2}$). It is posited that the accuracy will not be affected by the potentiometer offset error in a certain joint, provided that there exists a constant homogeneous transformation matrix (${}^{\hat{0}}T_0$) from the virtual base frame ($\hat{0}$) to the actual base frame ($0$), so that the virtual joint frame can be realigned to the actual joint frame despite the offset errors in all the preceding joints. The realignment is achieved by combining a constant transformation (${}^{\hat{0}}T_0$), with the result of hand-eye calibration (${}^{c}T_{\hat{0}}$) and any virtual transformations, \textit{e.g.} ${}^{\hat{0}}T_{\hat{1}}$.}

\begin{figure}[!h]
    \centering
    \includegraphics[width=0.75\columnwidth]{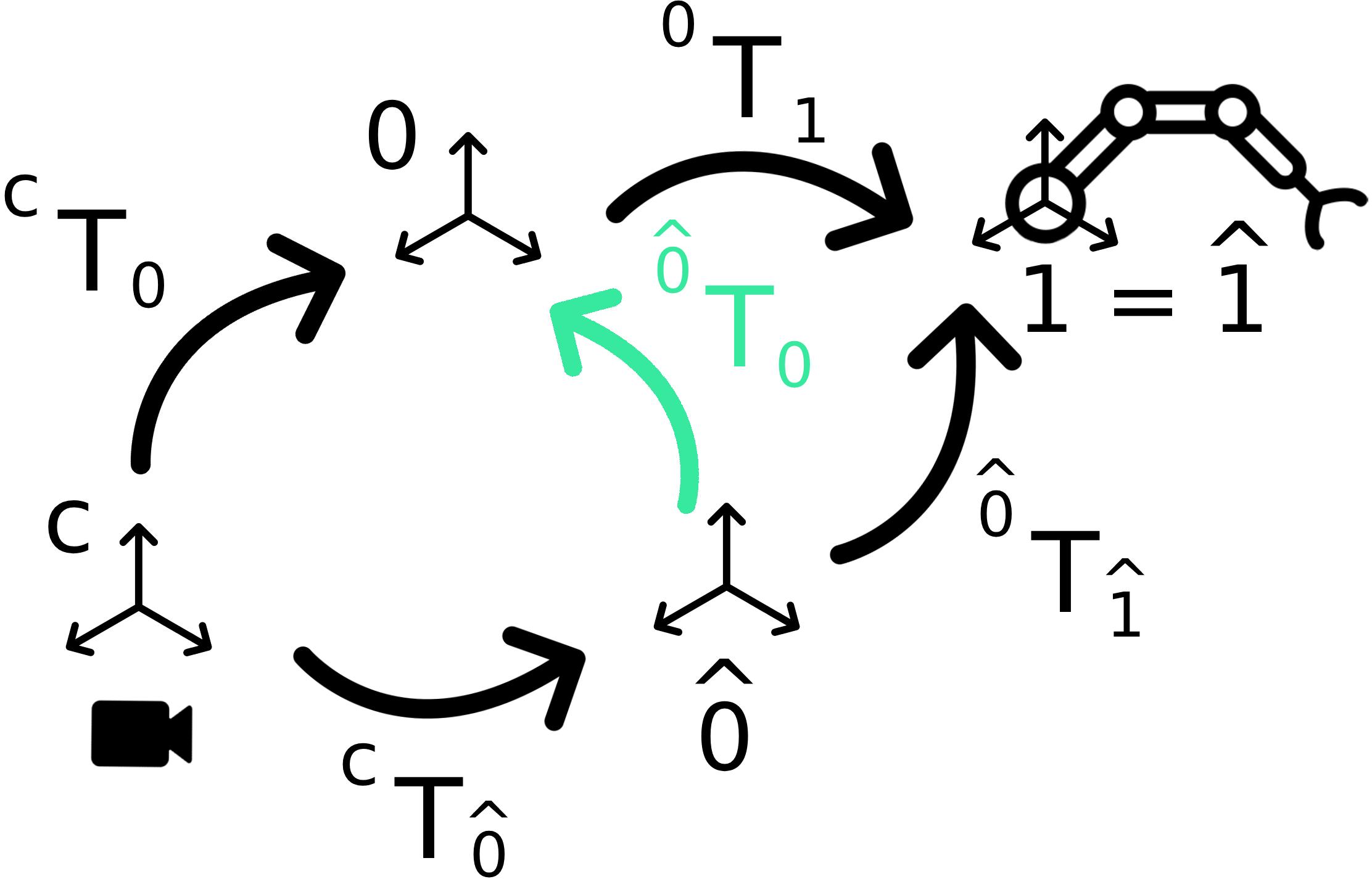}
    \caption{\add{This figure illustrates how hand-eye calibration compensates for the offset error in the first potentiometer with the transformation (${}^{c}T_{\hat{0}}$), from the camera frame ($c$) to the virtual base frame ($\hat{0}$). The new hand-eye calibration matrix ${}^{c}T_{\hat{0}}$ post-multiplied by ${}^{\hat{0}}T_{\hat{1}}$ results in the realignment for joint frame 1  ($1 = \hat{1}$), and hence the following joint frames will be aligned if no offset errors exist in the rest of the joints. Hand-eye calibration is capable of compensating for the potentiometer offset errors only when there exists a constant transformation matrix (${}^{\hat{0}}T_{0}$) from the virtual base frame ($\hat{0}$) to the actual base frame ($0$). It is later proved that hand-eye calibration only addresses the offset error in joint 1, but not in joint 2}}
    \label{fig:pot_app}
\end{figure}

\add{An example of how hand-eye calibration compensates for the offset error in the first joint is shown in Fig. \ref{fig:pot_app}. The goal is to find a constant transformation (${}^{\hat{0}}T_0$), so that after post-multiplying ${}^{\hat{0}}T_{\hat{1}}$ to the new hand-eye calibration matrix (${}^{c}T_{\hat{0}}$), realignment is achieved for joint 1 ($\hat{1} = 1$). After that, all subsequent joint coordinate frames will be aligned if no potentiometer offset errors exist in the remaining joints. For the first joint, the following equation is obtained:}

\begin{equation} \label{eq:offset1}
  {}^{\hat{0}}T_0 = {}^{\hat{0}}T_{\hat{1} = 1} ({}^{0}T_{1})^{-1}
\end{equation}

\add{The same logic also applies when the offset errors in both joint 1 and joint 2 are considered:}

\begin{equation} \label{eq:offset12}
  {}^{\hat{0}}T_0 = {}^{\hat{0}}T_{\hat{1}} {}^{\hat{1}}T_{\hat{2} = 2} ({}^{1}T_{2})^{-1}  ({}^{0}T_{1})^{-1}  
\end{equation}

\add{\noindent in which case, realignment is achieved for the second joint frame ($2 = \hat{2}$). Next, the analytical solution for ${}^{\hat{0}}T_0$ is found for the above three cases, with the modified DH transformation matrix being:}


\begin{equation*}
\resizebox{0.5\textwidth}{!}{$^{n-1}T_{n} = \begin{bmatrix}
    c(\theta_n) & -s(\theta_n) & 0 & a_{n-1} \\
    s(\theta_n) c(\alpha_{n-1}) & c(\theta_n) c(\alpha_{n-1}) & -s(\alpha_{n-1}) & -d_n s(\alpha_{n-1}) \\
    s(\theta_n) s(\alpha_{n-1}) & c(\theta_n) s(\alpha_{n-1}) & c(\alpha_{n-1}) & d_n c(\alpha_{n-1}) \\
    0 & 0 & 0 & 1
  \end{bmatrix}$}
\end{equation*}

\add{\noindent where $c()$ and $s()$ represent $sin()$ and $cos()$, respectively, and the DH parameters for the dVRK PSMs being:}

\begin{center}
\begin{tabular}{ |c|c|c|c|c| } 
 \hline
 $n$ & $\alpha_{n-1}$ & $a_{n-1}$ & $d_n$ & $\theta_n$ \\ \hline
 1 & $\frac{\pi}{2}$ & 0 & 0 & $q_1 + \frac{\pi}{2}$\\ 
 2 & $-\frac{\pi}{2}$ & 0 & 0 & $q_2 - \frac{\pi}{2}$\\
 \hline
\end{tabular}
\end{center}

\add{After solving Eq. \ref{eq:offset1} by substituting in the DH parameters above ($q_1 = j_1$ into ${}^{0}T_{1}$ and $q_1 = j_1 + \Delta_1$ into ${}^{\hat{0}}T_{\hat{1} = 1}$), the following constant transformation is obtained:}

\begin{equation*}
^{\hat{0}}T_{0} = \begin{bmatrix}
    c(\Delta_1) & 0 & -s(\Delta_1) & 0 \\
    0 & 1 & 0 & 0 \\
    s(\Delta_1) & 0 & c(\Delta_1) & 0 \\
    0 & 0 & 0 & 1
  \end{bmatrix}
\end{equation*}

\add{\noindent where $\Delta_1$ is the offset error in the first potentiometer (a constant value). This transformation represents a pure rotation around the y-axis of the actual base frame (0) by $-\Delta_1$.}

\add{Now, assume that $\Delta_1 = 0$ to single out the contribution of the offset error in the second potentiometer to $^{\hat{0}}T_{0}$ and solve Eq. \ref{eq:offset12} (substituting $q_2 = j_2 + \Delta_2$ into ${}^{\hat{1}}T_{\hat{2}=2}$ and $q_2 = j_2$ into ${}^{1}T_{2}$), the following transformation is obtained:}

\begin{equation*}
\resizebox{0.5\textwidth}{!}{$^{\hat{0}}T_{0} = \begin{bmatrix}
    c(j_1)^2+c(\Delta_2)s(j_1)^2 & s(\Delta_2)s(j_1) & -((-1+c(\Delta_2))c(j_1)s(j_1)) & 0 \\
    -s(\Delta_2)s(j_1) & c(\Delta_2) & c(j_1)s(\Delta_2) & 0 \\
    -((-1+c(\Delta_2))c(j_1)s(j_1)) & -c(j_1)s(\Delta_2) & c(\Delta_2)c(j_1)^2+s(j_1)^2 & 0 \\
    0 & 0 & 0 & 1
  \end{bmatrix}$}
\end{equation*}

\add{\noindent Notice that, when there exists an offset error in the second potentiometer ($\Delta_2$), $^{\hat{0}}T_{0}$ is not constant because it depends on the value of the first joint ($j_1$).}




\add{Therefore, it is concluded that a single hand-eye calibration only compensates for the offset error in the first potentiometer ($\Delta_1$) but not for the errors in the second potentiometer ($\Delta_2$). It also implies that, in the presence of potentiometer offset errors in joint 2, a non-constant hand-eye calibration matrix needs to be updated on the fly.}

\subsection{Evaluate potentiometer calibration accuracy}

The accuracy of potentiometer calibration can be evaluated by configuring a manipulator at multiple positions and examining if $\beta_E \approx \beta_P$ holds true for each joint across all joint configurations.
\section{Camera Calibration}
\label{sec:calib}

In this section, different aspects that need to be taken into consideration during the calibration of the endoscopic camera on the dVRK are discussed. It is assumed that a conventional computer vision method for camera calibration \cite{zhang2000flexible} is used, which relies on the pinhole camera model \cite{2011computervision} and a planar checkerboard with a known geometrical pattern.

\subsection{Need for accurate camera calibration}

Accurate camera calibration ensures the fidelity of algorithms that take endoscopic images as input for estimating 3D spatial information. Therefore, camera calibration plays a crucial role in image-based clinical applications such as image-guided surgery and autonomous robotic surgery, since these applications rely on algorithms that process 3D data. 
Such algorithms include hand-eye calibration, 3D reconstruction, 3D tracking, camera or instrument pose estimation, constraining the motion of surgical instruments to avoid collisions, interacting with the tissue via visual servoing, and registration of 3D structures in the endoscopic scene using Augmented Reality (AR). 

An image-based hand-eye calibration, for example, cannot be accurate unless the lens distortion and the camera intrinsic parameters are precisely calibrated. Similarly, a disparity-based 3D reconstruction cannot be accurate unless stereo images are accurately rectified, which requires not only accurate distortion and intrinsic parameters but also an accurate estimation of the geometric transformation from the left to the right stereo camera.

\add{\subsection{Challenges in camera calibration}}

\subsubsection{\add{Camera calibration parameters are not constant}}One fact that needs to be taken into consideration is that for the cameras of the dVRK, the calibration parameters should not be considered constant. These parameters might change unintentionally (due to camera self-heating \cite{yu2019modeling}, vibrations or mechanical shocks \cite{civera2009camera}), or intentionally by manual adjustment of the camera's focus, for example, to improve the imaging quality of an anatomical target or to adjust the camera's field-of-view. When the focus of a stereo camera is adjusted, its left and right camera sensors move relative to the tip of the endoscope. This sensor motion would cause changes in camera intrinsic parameters, principal point and lens distortion parameters \cite{pratt2014practical}, which hence requires the camera calibration to be repeated. It has been proposed to calibrate the camera before surgery \cite{mourgues2002flexible}; however, this is unrealistic because surgeons must first insert the endoscope inside the patient before adjusting the field-of-view or the focus to the target anatomy \cite{pratt2014practical}. Changing camera parameters is also an open problem for deep-learning image-based algorithms. The reason behind it is that a model that is trained on a dataset captured by one camera cannot be directly applied to a different camera. Even with the same camera, when a surgeon changes its focus setting, the model may become inaccurate because of changes in the camera parameters.

\subsubsection{\add{Human error during camera calibration}}
Common sources of error include inaccuracies of the calibration target (e.g. due to the target being printed on the wrong scale or being attached to a non-flat surface), the use of a small number of images or corners, limited distribution of the corners over the field of view of the camera, inaccurate 2D pixel positioning of the detected corners, overexposed corners due to a powerful endoscopic light source, and the presence of vibrations/motion when pictures are taken \add{\cite{bouguet2004camera, qi2010review}}.

\subsubsection{\add{The interlaced video problem}}
\begin{figure}
    \centering
    \includegraphics[width=0.7\columnwidth]{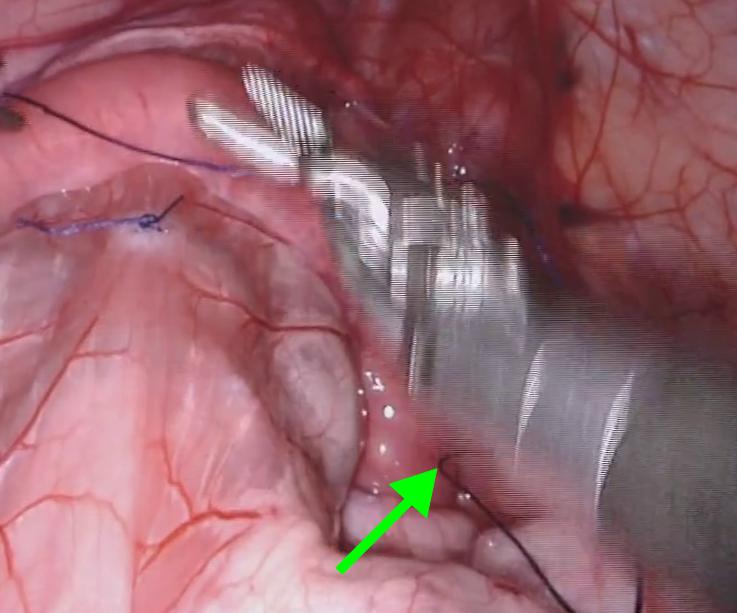}
    \caption{A visual example of the interlaced video problem on an image from an ex vivo surgery. This image was taken from the SurgT challenge \cite{surgt}.}
    \label{fig:interlaced_artifact}
\end{figure}

Interlaced video is a technique that creates an illusion of having a camera with a higher frame rate by doubling the frame rate (here from 25$Hz$ to 50$Hz$) while maintaining the same bandwidth \cite{zhao2021rethinking}. The downside of interlacing is that half of the image information at each timestamp is lost. The reason is that, in interlaced video, an image's pixels are captured at two sequential timestamps. At the first timestamp, the image's odd rows are captured and at the second timestamp, the even rows are captured, or vice-versa. Since the odd and even rows are captured at different timestamps, there can be significant artefacts when there is rapid movement from either tissue, the camera or surgical instruments. Figure \ref{fig:interlaced_artifact}, shows an example of this artefact in a video of a real \textit{ex vivo} surgery. \add{Interlaced video can have adverse effects on the accuracy of camera calibration. The reason is that, when the calibration target is in motion, it may be captured at two different poses, which are stored in the odd and even rows of a video frame. As a result, image keypoints can be distorted, which leads to inaccurate detection of keypoints in the calibration target.}

\add{\subsection{Methods for camera calibration}}

\subsubsection{Compensation for not constant calibration parameters}



In image-based robotic surgery, it is crucial that the robot is able to estimate the camera calibration parameters on-the-fly. Otherwise, the 3D information extracted from the endoscopic images will be inconsistent with the real dimensions of the surgical scene. Currently, there are three approaches to estimating the changeable camera calibration parameters during surgery, namely (a) using a calibration model, using (b) self-calibration, and (c) using a fixed-focus endoscope.

\subsubsection*{(a) Using a calibration model}

Pratt \textit{et al.} \cite{pratt2014practical}, created a calibration model which gathers the intrinsic and distortion parameters across the entire range of possible focal lengths of an endoscopic camera. The idea is that this model is first created preoperatively. Then during surgery, with the help of a single image of a calibration pattern attached to a surgical instrument, the focus setting can be estimated, and thereby its corresponding distortion and intrinsic parameters can be retrieved from the pre-created calibration model. The paper has also found that the geometric transformation from the left to the right camera is essentially constant for all the focus settings. 
One limitation of this paper is that it assumes that a calibration pattern is available intraoperatively. To address this drawback, Kalia \textit{et al.} \cite{kalia2019marker} made the assumption that given the fact that the focus is typically adjusted to the target anatomy, and that a surgical instrument is usually in contact with the target anatomy, the distance from the camera to the instrument can determine the focus setting. Then again, similar to Pratt \textit{et al.} \cite{pratt2014practical}, this focus setting is used to extract the corresponding parameters from the pre-created calibration model. \add{The second method presupposes that the surgical instrument's tip is in focus with the camera, which is not always the case. Moreover, it relies on accurate hand-eye calibration, as it involves estimating the instrument's pose relative to the camera. As a result, while it can be utilized in surgical procedures, the method's accuracy relies upon the soundness of these assumptions.}


\subsubsection*{(b) Using self-calibration}


In computer vision, self-calibration (or auto-calibration) estimates the camera calibration parameters using raw images without the requirement of a calibration pattern. Therefore, self-calibration methods \add{can be used intraoperatively}, as long as (1) these methods are able to estimate the calibration parameters on-the-fly, (2) are able to deal with non-rigid scenes, such as deforming soft-tissue. The degree of the deformation is proportional to the magnitude of the forces applied to the target anatomy. Self-calibration methods are mostly tailored for rigid world scenes. The earlier works solely focused on estimating the focal length \cite{azarbayejani1995recursive, qian2004bayesian}, but were later enhanced to estimate the focal length and principal point in addition to radial distortion \cite{civera2009camera, keivan2015online}. More recently, deep learning methods estimate all these parameters either via supervised learning \cite{bogdan2018deepcalib}, which trains with ground-truth calibration parameters, or via self-supervised learning \cite{gordon2019depth, fang2022self}, which trains directly from raw videos. In addition to research on rigid world scenes, methods for self-calibration on non-rigid surgical scenes have also been developed. Some of these methods estimate focal length only \cite{stoyanov2005laparoscope, bartoli2013robust, parashar2018self}. A more recent method estimates focal length and principal point in addition to radial distortion for non-rigid scenes \cite{agudo2021total}.

\subsubsection*{(c) Using a fixed-focus endoscope}

Taking the changeable calibration parameters problem into consideration, and given that adjusting the focus during surgery is impractical, the newer da Vinci systems are now using fixed-focus cameras \cite{davinci_new_endoscope}, also referred to as focus-free endoscopes. These new-generation endoscopes are still not available to dVRK systems and there is still little information about whether the calibration parameters on fixed-focus endoscopes are constant or not over time. \add{Despite the lack of information regarding the stability of calibration parameters over time for these endoscopes, they remain a top choice for clinical translation, even if they change slightly over time. This is because any small changes in calibration parameters could be corrected either preoperatively or intraoperatively using self-calibration techniques.}

\subsubsection{Compensation for human error during calibration}

If a checkerboard is used, \add{for laboratory experiments,} then it should be held at a static pose while capturing each of the camera calibration images. The reason is twofold. First, in the presence of checkerboard movement, inconsistent poses can be captured on the left and right stereo images if the cameras are not properly synchronized, and consequently, the estimation of the geometric transformation between two stereo cameras is inaccurate. The other reason is to avoid motion artefacts, which can significantly decrease the quality of camera calibration. Images containing motion artefacts should be pruned out of the calibration data, and fortunately, these motion artefacts can easily be identified thanks to the interlaced video technology that the dVRK endoscope camera is currently using. \add{It is recommended to inspect the quality of the detected corners in the checkerboard images. This can be done by checking if the corners are correctly positioned between the squares of the checkerboard. It is also recommended to inspect the overall reprojection errors, which can be used to identify possible outliers \cite{li2013multiple}.} \add{A common approach for verifying lens distortion parameters is to check if straight lines in the real-world are projected into straight lines in the 2D image \cite{hartley2003multiple}. Another technique for verifying stereo extrinsic and intrinsic parameters involves checking if keypoints are row-aligned in the rectified images when the left and right images are stacked horizontally \cite{hartley2003multiple}.}

\subsubsection{Compensation for interlaced video}

Many methods have been proposed for intra-frame deinterlacing.
A popular solution for keeping the original resolution and perform real-time removal of this artefact is to use every other row (i.e., either the even rows or the odd rows) and interpolate the pixels for missing rows. 
There are other more complex algorithms for deinterlacing, such as recent deep learning-based approaches \cite{liu2021spatial, yeh2022vdnet}. Note that no matter how complex the algorithm is, the loss of image information at any given timestamp is inevitable. \add{To avoid any possible impact of interlaced video on camera calibration, best practice is to maintain a static position for the calibration target during image capture. By doing so, the complete raw image from the dVRK can be utilized directly without requiring any deinterlacing. Moreover, holding the target stationary ensures that there is no motion blur, and any potential synchronization discrepancies between the left and right cameras do not affect the calibration precision. Nevertheless, it should be noted that maintaining a static position for the calibration target during image capture is only feasible in a controlled laboratory setting. In real-world scenarios, such as during intraoperative camera self-calibration, interlaced video remains a concern as it can potentially affect calibration precision. Therefore, deinterlacing may be necessary to utilize the complete raw image from the dVRK for self-calibration methods.}
\section{Hand-eye calibration}
\label{sec:Hand-eye}
Hand-eye calibration has been widely researched in the area of robotics; it aims to find the transformation from the camera frame to the robot frame. There exists some ambiguity in the definition of a hand-eye calibration problem in the dVRK community, with some researchers defining it as finding the transformation from the camera frame to the frame of the robot arm to which the camera is attached \cite{Allen_max_detailed_handeye,7759387}, while most other researchers place their focus on finding the transformation from the camera frame to the base frame of another robot arm to which a surgical tool is attached. Although for a generalized hand-eye calibration problem, the robot frame is unspecified, in the remaining part of this section, we define the dVRK hand-eye calibration problem as per the latter school of thought. \add{Hand-eye calibration is related to both kinematics calibration (including potentiometer calibration) and camera calibration. Therefore, it inherits the inaccuracy discussed in the previous sections.}

\subsection{Need for hand-eye calibration}
The importance of developing an accurate and time-efficient hand-eye calibration strategy is twofold. In a teleoperation scenario, an accurate hand-eye transformation matrix enables a better alignment between the surgeon's view and the endoscopic camera view so that surgeons can see the surgical tools move in sync with their hand movements, making the teleoperation more intuitive. Through an operation, there are times when the robot arms need to be repositioned to adjust to different surgical tasks, and whenever its ``set up joints" (SUJ) are relocated, the hand-eye calibration needs to be recomputed. Despite the fact that it is desirable to develop a hand-eye calibration method that is both accurate and time-efficient, in a teleoperation scenario, time-efficiency takes priority over accuracy because humans can compensate for a slight misalignment between the movements of their hands and the surgical tools. However, for automatic surgical tasks, the accuracy of hand-eye calibration is of paramount importance because all robot motion commands are generated out of visual information captured by the endoscopic camera, and these visual cues cannot be correctly perceived by the robot arm unless an accurate transformation matrix is provided. 

\subsection{Challenges in dVRK hand-eye calibration}
Although there is a rich literature on hand-eye calibration problems in a generalized robotic framework, the development of calibration strategies for the da Vinci robot, which is customized for MIS, is still challenging due to several intrinsic constraints of the system. As observed in \cite{handEye_review,BIT_handEye_review,handEye_quaternion,tsai_calibration}, most widely-adopted hand-eye calibration methods require an external calibration object such as a planar checkerboard and external hardware. However, these methods are not applicable in a MIS environment where the working space is narrow and introducing external calibration objects is impractical. In addition, the Remote Center of Motion (RCM) constraint \cite{JHU_handeye_calib} of the patient-side robots limit the number of possible joint configurations available, which renders general calibration methods yet more challenging to implement in this context. 

\subsection{dVRK hand-eye calibration methods}
Most hand-eye calibration problems pivot on solving $AX=XB$ \cite{AX=XB_1987} or its derivative $AX=YB$ \cite{AX=YB_1994}, where $X$ represents the transformation matrix of interest, and $A$ and $B$ are matrices that can be deduced from kinematic readings and extrinsic calibration, such as the checkerboard calibration method \cite{checkerboard_pnp}. To allow for the specific constraints of the da Vinci robot, researchers in the dVRK community build upon traditional hand-eye calibration methods to produce bespoke approaches that are better suited for the task. Although there are a handful of research works that rely on classical calibration objects (e.g. \cite{QR_code_hand_eye_calib,NDI_polaris_hand_eye_calib}), most recent approaches have opted for more nuanced implementations. Pachtrachai \textit{et al.} \cite{7759387} proposed an approach which utilizes a tool-tracking algorithm so that the surgical tool itself functions as a calibration object \textit{per se}. Cartucho \textit{et al.} \cite{cartucho_self, cylinder_marker} designed a cylindrical marker that can be attached around the shaft of a surgical tool, serving as a compatible calibration object. And Wang \textit{et al.} \cite{JHU_handeye_calib} developed an algorithm which purely depends on the dVRK kinematic readings and image information captured by the endoscopic camera. Much as great progress has been made, the benefit of avoiding the use of external calibration objects has come at the price of worse calibration accuracy. 

To improve the calibration accuracy and the robustness of calibration algorithms against noise, research also focuses on designing advanced $AX=XB$ solvers \cite{Allen_max_detailed_handeye,7759387,zhangLin_handeyecalib}. In cases where the hand-eye calibration needs to be performed in real-time to compensate for the dVRK kinematic positioning error, computational efficiency must also be considered \cite{zhangLin_handeyecalib}. Zhong \textit{et al.} \cite{Hand_eye_nonAX_method} also proposed an Interactive Manipulation-based (IM) computation method that enables fast extraction of the position and orientation term of hand-eye calibration, although they did not regard hand-eye calibration as $AX=XB$ or $AX=YB$ problem in their work.
\section{Discussion}
\label{sec:discussion}

\add{The data information provided by the first generation dVRK is not flawless. Hence, without a proper understanding of the system's constraints and calibration processes, the performance of surgical tasks can fall short of expectations.} \add{Due to page limits, this paper did not cover other technical limitations, such as inappropriate controller gains etc., which could also serve as valuable research directions for the dVRK community.}

\begin{table*}[]
    \caption{This table provides guidance on the implementation of each calibration based on different research objectives. `$\checkmark$' represents ``Usually required" and `$-$' represents ``Usually not required".}
    \centering
    \begin{tabular}{|l|c|c|c|c|c}
      \hline \multirow{2}{*}{Requirements}
       &  \textit{Kinematics} & \textit{Potentiometer} & \textit{Camera}     & \textit{Hand-eye} \\ 
                 &  \textit{calibration}     & \textit{calibration} & \textit{calibration} & \textit{calibration}  \\
      \hline
      Human teleoperation? & \multirow{2}{*}{-} & \multirow{2}{*}{-} & \multirow{2}{*}{-} & \multirow{2}{*}{-} \\ (e.g. surgeon training, skill assessment and gesture recognition) & & & &\\ \hline Requires accurate encoder position readings?  & \multirow{2}{*}{\checkmark} & \multirow{2}{*}{\checkmark} & \multirow{2}{*}{-}  & \multirow{2}{*}{-} \\ (e.g. System simulation and modelling) & & & & \\ \hline Requires extracting 3D information from images? & \multirow{2}{*}{-} & \multirow{2}{*}{-} & \multirow{2}{*}{\checkmark}  & \multirow{2}{*}{-} \\ (e.g. Augmented Reality, Depth Estimation, 3D Tracking) & & & & \\ \hline Requires accurate end-effector pose relative to the camera? & \multirow{2}{*}{\checkmark} & \multirow{2}{*}{\checkmark} & \multirow{2}{*}{\checkmark}  & \multirow{2}{*}{\checkmark} \\ (e.g. Autonomous Control) & & & & \\ \hline
    \end{tabular}
    \label{tab:calibs}
\end{table*}

\add{Considering the diverse research interests in the dVRK community, we aim to provide general guidance for dVRK users about the implementation of these calibrations based on the repercussions of these implementations on their research outcomes, as shown in Table.\ref{tab:calibs}. First, the need to conduct specific calibrations should pivot on the type of data being used, these data spanning image data, kinematics data, dynamics data etc. From this perspective, it is the researchers who need to be aware of the defect in these data and conduct corresponding calibrations. For example, in automating surgical subtasks, it is a common practice to use both image and kinematics data for localising the target position and commanding a surgical instrument to move towards the target, and therefore all of the four calibrations are needed to achieve an optimal performance. In studies involving AC, where accurate tip position is needed for enquiring the minimum distance between a surgical instrument and a moving tissue surface, it is necessary to perform kinematics calibration. Second, we suggest that the existence of a human element in a study largely diminishes the importance of conducting these calibrations, because a human can compensate for inaccuracies. Therefore, for researches that feature surgeon training, teleoperation assistance, where the accuracy requirement is less stringent, the implementation of these calibrations are left to the discretion of the researchers themselves.}

Kinematics calibration aims to reduce the dVRK positioning error. We have narrowed error factors down to three categories, namely ``cable-driven effects", ``inaccurate kinematics parameters" and ``other non-kinematic sources". \add{In this paper, we evaluate these calibration methods based on their application scenarios, given that some are designed for laboratory trials while others are for clinical use. Overall, the methods that rely on an external marker or a calibration object are not suitable for clinical application. Although the calibration methods that build upon a pre-training model have the potential to be translated into clinical practice, as they are implemented preoperatively, whether the effectiveness of these calibration methods will diminish over time remains to be assessed. Therefore, it is preferable to develop intraoperative calibration methods that take real-time information into consideration.}

Potentiometer calibration is crucial to obtain accurate joint encoder readings, thereby indirectly impacting the accuracy of tip-pose estimation. \add{Despite this, until now, the dVRK community had not yet investigated the inherent nonlinearities of potentiometers. One of the challenges that have been addressed is the calibration of the potentiometer offsets. These offsets are difficult to calibrate as their calibration requires users to position the actuators at their zero-position. For the PSMs, techniques have been developed to calibrate the offset of the last five potentiometers. However, calibrating the offset for the first two potentiometers remains an open problem. It is discussed in this paper that the first potentiometer offset error can be compensated for using hand-eye calibration, but this does not apply to the second potentiometer. Additionally, it has been identified that each 1\textdegree error on the second joint encoder reading results in a mean error of 3.4 \textit{mm} in the end-effector position. Therefore, it is essential to develop a technique for calibrating the offset of the second potentiometer, which is still an open problem for the community. Furthermore, for the ECM and SSMs, there is still no technique to determine their exact zero-position for offset calibration.}


Regarding camera calibration, obtaining accurate values of lens distortion, camera intrinsic parameters and the transformation between two stereo cameras is crucial for tasks that extract 3D information out of endoscopic images.
With current endoscopic technology employed with the dVRK, calibrated parameters are not constant, mainly due to the dynamic adjustment of focus settings, which happens frequently in real surgeries. \add{In the authors' opinion, it would be preferable if dVRK researchers focus on developing methods to continuously update current camera parameters using natural surgical video without the aid of a calibration target that is unusable in real surgery. However, for laboratory experiments, a calibration target such as a checkerboard can be used. Furthermore, another interesting observation is that instead of trying to deinterlace input images for deep learning models, one can choose to use only every other row (e.g., even rows). This is because most deep learning models currently require lower resolution images, and therefore one can use only half of the rows (without interlaced video issues) as a natural down-scaling technique.}

Hand-eye calibration has been widely researched in the robotics area, but most of these developed methods are not suitable for an RMIS setting, where the task space is narrowly confined, and the accuracy requirement can be much higher. \add{In a clinical setting, the da Vinci robots are already equipped with sensorized SUJs, which enable automatic updates of hand-eye transform. Although the inaccuracy is up to around 5 $cm$ \cite{dvrk_wiki}, it suffices for teleoperation scenarios due to human compensation. From this perspective, all current works focusing on the development of hand-eye calibration with high accuracy requirements are to be applied for laboratory trials. In theory, hand-eye calibration only needs to be performed once before experiments. However, due to the nature of hand-eye calibration resolving the inaccuracy in encoder readings, it would be more efficient to conduct on-the-fly hand-eye calibration for applications where accurate kinematics data is also required. Furthermore, considering that hand-eye calibration mostly benefits surgical applications where visual guidance plays a crucial role, we suspect that more emphasis will be placed on the development of clinically adaptable hand-eye calibration methods with high accuracy requirements once the need for greater uptake in surgical subtasks automation will have materialized.} 

\add{We believe that, as newer versions of the dVRKs gradually become available, many of the technical issues under the first-generation dVRKs discussed in this paper can be alleviated to a large extent, if not fully eradicated. For example, the latest da Vinci \RomanNumeralCaps{11} robots adopt digital potentiometers, as opposed to analogue ones installed on the da Vinci Classic, and hence the accuracy and reliability of joint encoder readings could be potentially enhanced for future versions of the dVRK. Also, since the advent of the da Vinci Si, the endoscope cameras installed on the robots produce images with better quality, which would inherently improve the quality of visual feedback for future dVRK researchers.}


\add{Nevertheless, it is likely that future da Vinci robots will remain tendon-driven despite having shorter kinematics chains; therefore, essential kinematics calibration will still be required. In addition, the practice of moving the SUJs and zooming in/out the endoscopic camera during operation will still require repeating hand-eye calibration and camera calibration. To streamline the overall surgical workflow, we suggest that one avenue for future research should be the development of a unified approach that could merge the calibration procedures and be implemented on the fly. We have stressed the interrelationship of kinematics calibration, hand-eye calibration and potentiometer calibration, which supports the benefit of a unified approach. We envision that a single clinically adaptable marker, attached around the tool shaft, could serve as a mutual calibration object, with great potential to unify the calibration procedures. The unified approach would help reduce potential error factors from individual research efforts by future dVRK users.}

\section{Conclusion}
In this paper we covered the most widely-reported dVRK technical limitations discovered in the currently available literature with a view to aiding future research on the subject. However, there are other underlying technical constraints that may have equal, if not more of a severe impact on surgical task performance. These underlying constraints, including but not limited to misalignment between surgeon-side and patient-side manipulators and naive inverse kinematics solvers, have already been discovered by leading researchers of the dVRK community. This paper not only aims to provide guidance for dVRK users such that they are aware of these potential factors, but also to encourage future works on the comparative assessment of different calibration and performance accuracy assessment studies. The constraints identified here are rooted in the first-generation dVRK teleoperational surgical platform, though are likely to exist in similar form as new versions are made available. Therefore, the development of unified solutions that can be incorporated within future versions of da Vinci robots are also recommended. Finally, given that the cable-driven nature and imaging system of the da Vinci system are mirrored in other surgical robotic implementations (e.g. RAVEN), the hope is that our paper can also be of use by researchers outside the dVRK community.
\section{Acknowledgments}
We would like to express our appreciation towards \add{three} leading researchers in the dVRK community: Mr. Anton Deguet\add{,} Dr. Simon DiMaio \add{and Prof. Peter Kazanzides} for their valuable advice during the course of completing this work, especially to Anton, who 
kindly agreed to review the potentiometer calibration section of this paper, \add{and Peter, who provided valuable feedback towards our revised manuscript}. Also, we thank each and every one of the peer researchers within the dVRK community who completed the questionnaire and provided us with invaluable feedback based on their unique research experience. 

\appendices
\section{Questionnaire}
\label{sec:questionnaire}

The questionnaire consists of three sections, corresponding to kinematic/dynamics data (data from the PSMs, ECM, and SSMs), imaging data (images captured by the dVRK's endoscope), and system data (e.g., signals from the foot pedals for teleoperation), respectively. For each section, users first need to answer a yes-or-no question to determine if they have used the type of data before. If not, they could skip to the next section; otherwise, they are instructed to point out the sources of error in that data type which they reckon may have affected their research results. Users are prompted to identify the sources of error from a list of selected options that were gathered from literature review and consultation with leading researchers. In addition, the participants are also available to put in other potential sources of error by themselves. A total of 35 responses have been collected from 20 different institutions, summarized as follows:

\subsection{Kinematics and dynamics data}

\begin{figure*}[t]
    \centering
    \includegraphics[width=0.95\textwidth]{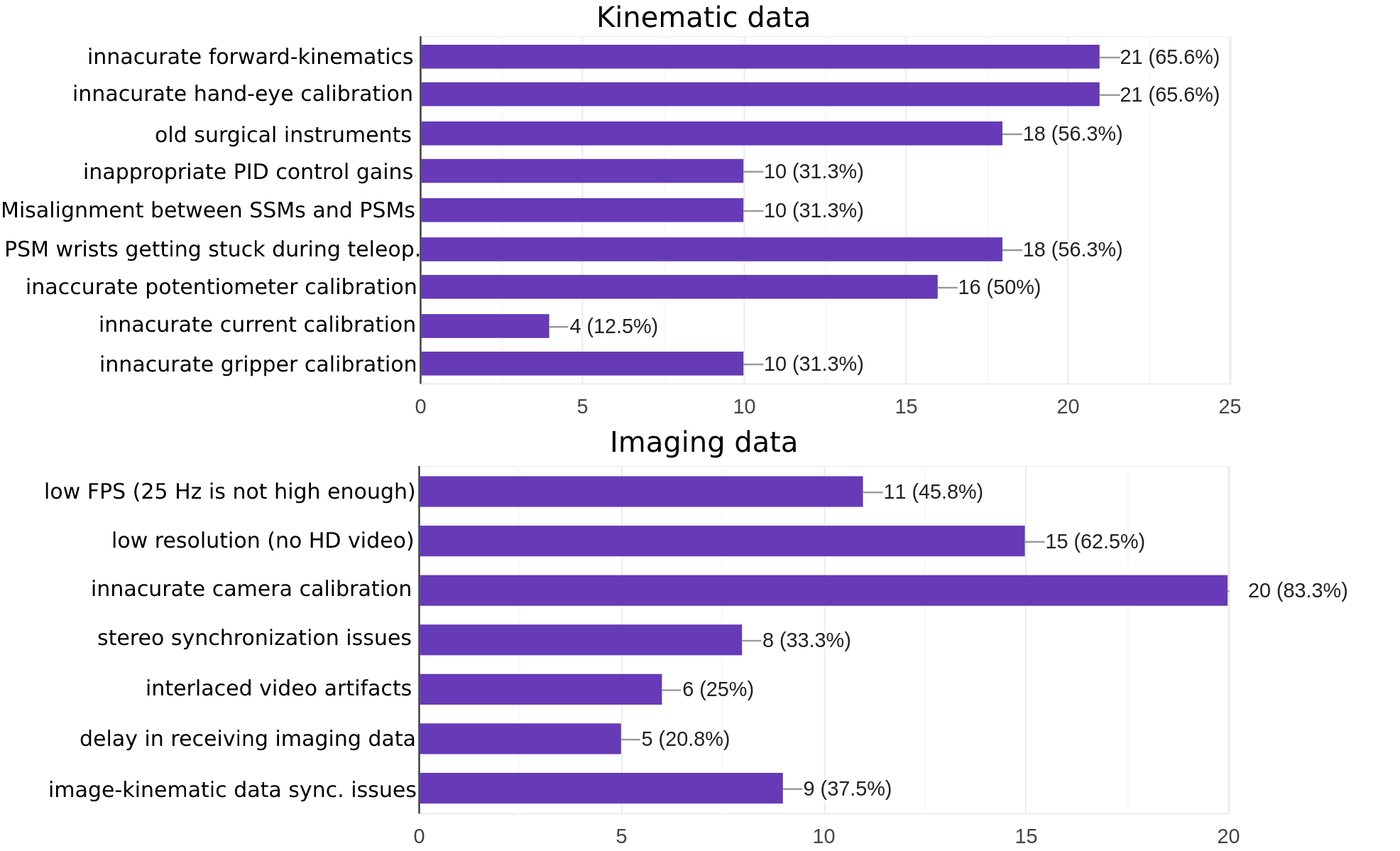}
    \caption{Answers to the questionnaire identifying sources of error both on kinematic and imaging data captured by the first-generation dVRK.}
    \label{fig:questionnaire}
\end{figure*}

Approximately 95\% of the participants have experience in dealing with kinematics and dynamics data. The most commonly identified problems are ``inaccurate forward-kinematics" (66\%) and ``inaccurate hand-eye calibration" (66\%), followed by issues with the surgical instruments (56\%), and ``inaccurate potentiometer calibration'' (50\%), as shown in Figure \ref{fig:questionnaire}.

\subsection{Imaging data}

Approximately 70\% of the participants have experience in dealing with imaging data streamed from the endoscope of the dVRK. The most commonly identified problem with imaging data is ``inaccurate camera calibration'' (83\%), followed by ``low pixel resolution'' (65\%), as shown in Figure \ref{fig:questionnaire}.

In addition to the sources of error listed from the selection box, 20\% of the participants have reflected that there is a significant difference between the left and right image, mainly perceived as colour mismatch. Also, 13\% of the participants have pointed out the presence of noise in the images.

\subsection{System state data}

Approximately 30\% of the participants have experience in dealing with system state data. No significant sources of error have been identified here.

\add{\subsection{Questionnaire Analysis}}

\add{After further analysis, it is discovered that 80\% of dVRK users did not conduct kinematics calibration when conducting hand-eye calibration despite they are in the knowledge that the accuracy of joint encoder readings could have been improved. We hypothesise that this results from a lack of readily available resources for kinematics calibration, which could be leveraged by dVRK users when undertaking their individual research.
Also, 90\% of researchers adopted marker-based approaches for hand-eye calibration, instead of marker-free ones. 
}

\add{It is also discovered that a significant percentage of dVRK users are either not aware of the necessity for certain calibrations, or not familiar with the methods for certain calibrations. Specifically, 40\% of the users are not aware that the camera intrinsic parameters are not constant; 20\% are not aware of the methods for kinematics calibration, and 30\% are not aware of potentiometer calibration. In addition, 25\% of the users reported that they bypassed hand-eye calibration in their research, which can be explained by the fact that some institutes mainly use the dVRKs for the purpose of training or conducting user studies, where the human element makes redundant the hand-eye calibration.
}

\add{After discussions with the participants, it is gathered that 63\% of the dVRK users are dissatisfied with the current standard definition (SD) cameras on the first generation dVRK. It is suggested from our analysis that the incorporation of cameras with high-definition (HD) and a faster frame rate into the dVRK system would be preferable, such that the image data would be better synchronized with the kinematics data, thereby contributing towards better system functionality and usability.
}


\addtolength{\textheight}{-0cm}   


\bibliographystyle{ieeetr}
\bibliography{IEEEabrv,files/references} 

\end{document}